\documentclass[10pt,twocolumn,letterpaper]{article}

\usepackage{iccv}              
\usepackage{algorithm}
\usepackage{algpseudocode}
\usepackage{multirow}
\definecolor{iccvblue}{rgb}{0.21,0.49,0.74}
\usepackage[pagebackref,breaklinks,colorlinks,allcolors=iccvblue]{hyperref}

\def\paperID{3538} 
\def\confName{ICCV}
\def\confYear{2025}

\title{Imbalance in Balance: Online Concept Balancing in Generation Models}

\author{
Yukai Shi$^{1,2*}$ ~\qquad
Jiarong Ou$^{2}$~\qquad
Rui Chen$^{2}$~\qquad
Haotian Yang$^{2}$~\qquad
Jiahao Wang$^{2}$~\qquad\\
Xin Tao$^{2}$$^{\dag}$~\qquad
Pengfei Wan$^{2}$~\qquad
Di Zhang$^{2}$~\qquad
Kun Gai$^{2}$~\qquad\\
$^1$~Tsinghua University~\qquad
$^2$~Kling Team, Kuaishou Technology
\\
}

\begin{document}

\twocolumn[{%
    \renewcommand\twocolumn[1][]{#1}%
    \maketitle
    \vspace{-3.5em}
    \begin{center}
    \includegraphics[width=\textwidth]{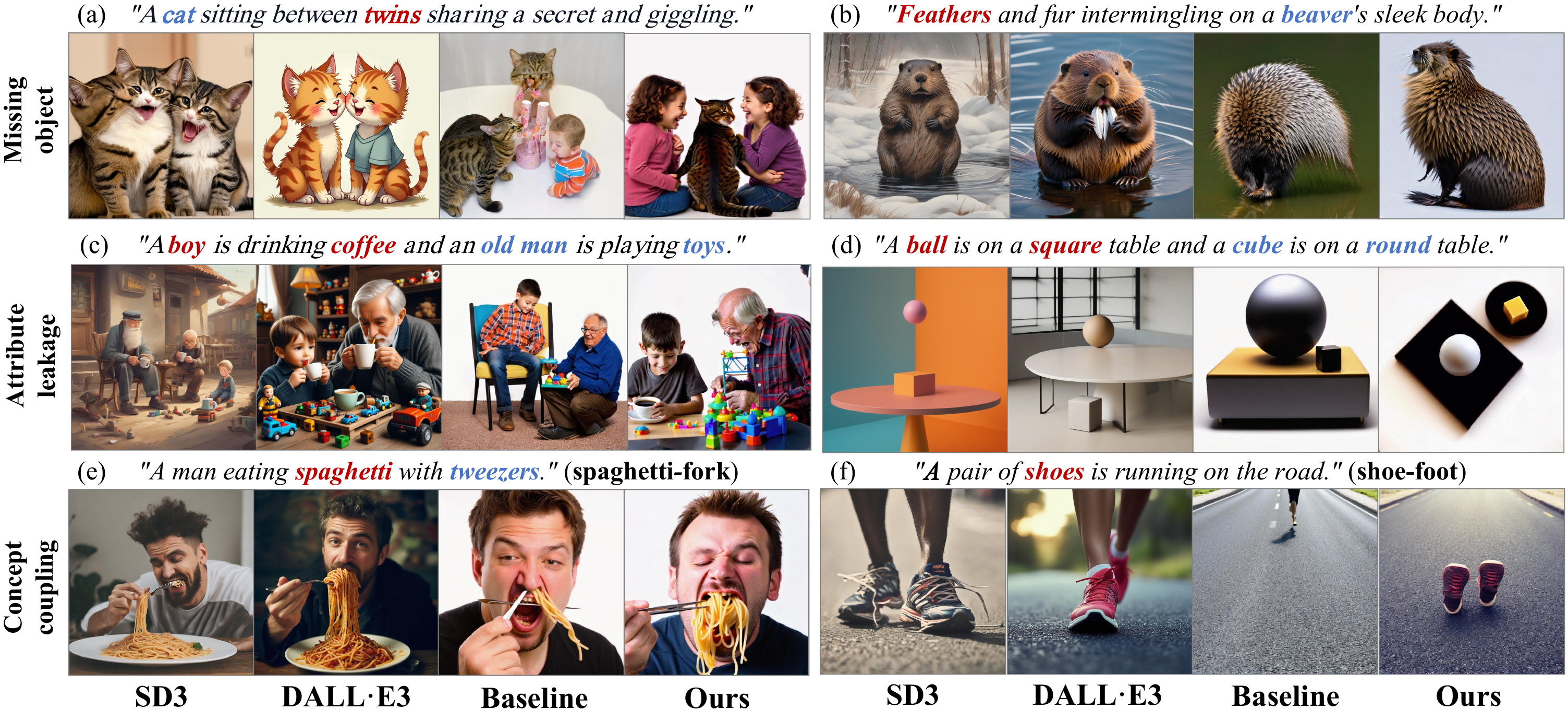}
    \end{center}
    \vspace{-1.5em}
    \captionof{figure}{\textbf{Our method achieves better concept composition ability with much smaller dataset (31M).} Existing models face missing object, attribute leakage, and concept entanglement problem. Specifically, Figure(a)(b) miss the expected concepts (twins, feather). Figure(c)(d) incorrectly match the attribute of the subjects. Figure(e)(f) exists unnecessary concepts (fork, legs).}
    \vspace{1em}
    \label{fig:teaser}
}]

\renewcommand{\thefootnote}{\fnsymbol{footnote}}
\footnotetext[0]{*Done during internship at Kling Team. $^{\dag}$ Corresponding author.}

\begin{abstract}
In visual generation tasks, the responses and combinations of complex concepts often lack stability and are error-prone, which remains an under-explored area. In this paper, we attempt to explore the causal factors for poor concept responses through elaborately designed experiments. We also design a concept-wise equalization loss function (IMBA loss) to address this issue. Our proposed method is online, eliminating the need for offline dataset processing, and requires minimal code changes. In our newly proposed complex concept benchmark Inert-CompBench and two other public test sets, our method significantly enhances the concept response capability of baseline models and yields highly competitive results with only a few codes released at \url{https://github.com/KwaiVGI/IMBA-Loss}.
\end{abstract}    
\section{Introduction}
\label{sec:intro}
\vspace{-0.5em}
In recent years, visual generative models have witnessed remarkable progress, attributed to the improvements of generation paradigms (\eg diffusion models ~\citep{song2020score,lipman2022flow}, auto-regressive models~\citep{mar,var,lee2022autoregressive,sun2024autoregressivebeats,titok}), the proposal of new network architectures (\eg DiT~\citep{dit,sd3,chen2023pixart}, \etc), and the release of large-scale dataset (\eg~\citep{laion,chen2024panda,wang2024koala}, \etc).  Driven by the concerted efforts of academia, industry, and the community, these models have reached a high level of sophistication. They are now capable of generating strikingly realistic images (\citep{imagen,latentdiffusion,podell2023sdxl,dalle3,midjourney,sd3}), videos (\citep{kling,imagenvideo,snapvideo,sora,latte}), and 3D models (\citep{li2024craftsman,zhang2024clay,wu2024direct3d}), and have been extensively adopted in creative endeavors and content generation scenarios. Remarkably, the most attractive aspect of generative models is their proficiency in deciphering and recombining real-world concepts. This enables them to conjure up objects and scenes that do not even exist in the physical world. This distinctive capability significantly expands the creative boundaries for content producers, unlocking novel opportunities for innovation.

However, in real-world applications, generative models often struggle to consistently generate outputs that closely match user expectations, especially in the realm of concept composition. Consider text-to-image (T2I) generation: even the most advanced models~\citep{dalle3,sd3,midjourney,flux} often suffer from problems like \emph{concept missing, attribute leakage} and \emph{concept coupling} as shown in Figure~\ref{fig:teaser}. To address these issues, several training-free approaches \citep{chefer2023attend,liu2022compositional,wang2023compositional,chen2024traininglayout,feng2022trainingstructure} have been proposed. Although these methods can be effective in specific scenarios, we contend that several essential problems underlying this challenge still remain under-explored:

\vspace{-1em}
\paragraph{Causal Factors} Visual generation models are complex systems with multiple components, and their performance is affected by many factors. Previous efforts \citep{okawa2023compositionalemerge,wiedemer2023compositionalprinceples,zhao2024lost} have yielded interesting insights, often using simple synthetic data (\eg~basic shapes of different sizes and colors) or focusing on class-to-image (C2I) tasks. However, these data are too simplistic to mirror the true complexity of text-to-image (T2I) tasks, where the number of concepts and their combinations far exceed those studied before. To bridge this gap, we are the first to conduct in-depth analysis on large-scale text-image pair data. We propose and verify the following hypotheses one by one: 1) When the dataset reaches a sufficient scale, concepts will naturally be covered and learned comprehensively. 2) As the model size increases, it becomes easier for the model to learn concept responses effectively. 3) The distribution of the dataset itself plays a dominant role. Through elaborately designed experiments, we have uncovered some intriguing findings. Firstly, an increase in dataset scale does not lead to improved responses for combined concepts. Secondly, once the model size reaches a certain threshold, there is no further enhancement in complex concept responses. Thirdly, a more balanced data distribution can significantly boost the model's ability to respond to combined concepts.

\vspace{-1em}
\paragraph{The Panacea} Based on previous analysis, our work, like prior studies \citep{leevy2018surveyimbalance,johnson2019surveyclassimbalance}, focuses on optimizing data distribution. However, this endeavor is rife with challenges. First, the large data scale in text-to-image (T2I) tasks renders any preprocessing for dataset-level statistical analysis and equalization~\citep{yan2024trainingclassimbalance,katsumata2024labelaug} prohibitively costly. Second, since each text prompt typically contains multiple concepts, sample equalization using loss weights \citep{cui2019classloss,park2021influence} has limited effectiveness in balancing concept distributions. Third, given the diverse generation paradigms and high model training costs, an ideal solution should be plug-and-play and applicable across different models.

In this paper, we introduce a \textbf{concept-wise} equalization approach. First, we acknowledge the \textbf{im}balance within the seemingly \textbf{ba}lanced dataset, where the distribution of concepts in the training data is uneven. By taking this into account, we ingeniously approximate the ideal balanced distribution using IMBA distance, which effectively captures the data distribution with unconditional generation results. Then, we develop a token-wise reweighting strategy (\emph{a.k.a} IMBA loss) for training. Our method is simple: it only requires a few lines of code to modify the loss function, without significant changes to the training process, and is compatible with various diffusion models.

\vspace{-1em}
\paragraph{Evaluation} Unlike general image generation benchmarks, the ability to combine concepts can only be differentiated on more complex and targeted test sets. In addition to evaluating on existing benchmarks such as T2I-CompBench~\citep{huang2023t2icompbench} and LC-Mis benchmark~\citep{zhao2024lost}, we have carefully constructed a new benchmark called Inert-CompBench. We identify inert concepts (difficult to integrate with other concepts) from large-scale text-image datasets and combine them with head concepts to obtain caption candidates. Then we construct a concept graph and filter out the uncommon-sense compliant concept pairs based edge weight. Finally, we generate 5 captions for each pairs with LLM~\citep{achiam2023gpt} for evaluation.

In summary, our contributions are as threefold:
\begin{itemize}
    \item We illustrate that once the model and training data attains a substantial scale, data distribution becomes the primary determinant of the model's concept composition ability.
    \item We propose the concept-wise equalization approach (IMBA loss) to address imbalanced concept distribution in training data. It is easy to implement, cost-effective, and applicable to different models. Promising results were obtained on three benchmarks.
    \item We introduce a novel concept composition benchmark named Inert-CompBench. This benchmark encompasses concepts that pose challenges for composition in an open-world scenario, complementing existing benchmarks.
\end{itemize}

\section{Related Work}
\label{sec:related works}

\subsection{Concept Composition}
Concept composition is the ability of generative models to accurately generate content with multiple concepts, reflecting their learning and understanding capabilities. It is a crucial indicator of a model's generalization ability. In text-image generation, several benchmarks~\citep{chefer2023attend,huang2023t2icompbench,zhao2024lost} have been proposed to comprehensively evaluate this ability. However, as shown in Figure~\ref{fig:teaser}, existing pre-trained models~\citep{sd3,midjourney,dalle3,podell2023sdxl} still suffer from issues like missing object, attribute leakage and concept entanglement. Some studies~\citep{okawa2023compositionalemerge,chang2024skews,wiedemer2023compositionalprinceples} based on synthetic experiments find that the concept composition ability of diffusion models is related to data completeness, balance, and disentanglement. However, these studies are often small-scale and class-conditioned, creating a gap with text-image tasks. From an application perspective, some works~\citep{chefer2023attend,feng2022trainingstructure,liu2022compositional,wang2023compositional} propose training-free methods by optimizing attention maps to enhance the model's concept response strength, when others~\citep{manas2024improving,ding2024freecustom} add input modalities or generating through multiple rounds of feedback. These methods are often limited by the capabilities of the foundation model. And there is little work analyzing the factors determining concept composition ability from a pre-training perspective. Our work aims to address these gaps.

\subsection{Data Balancing}
Data balancing aims at learning generalized models from long-tailed data distributions. Many works~\citep{zhang2023deep,leevy2018surveyimbalance,johnson2019surveyclassimbalance} have achieved excellent results in class-specific tasks, such as re-sampling, re-margining, and re-weighting~\citep{lin2017focal,cui2019classloss,park2021influence,ren2020balanced}. Some studies~\citep{qin2023classbalancing,yan2024trainingclassimbalance,um2024self,zhang2024long,katsumata2024labelaug} have extended class-based re-weighting to class-image generation tasks with inter-class distance, balanced distribution, label augmentation, or self-guided methods, showing impressive results. However, since text prompts is a joint distribution of multiple classes and each image cannot be assigned a single class, data balancing in text-image tasks still needs further exploration.
\section{Causal Factors of Concept Composition}
\label{sec: analysis}

In this section, we bridge the gap between synthetic experiments~\citep{okawa2023compositionalemerge,wiedemer2023compositionalprinceples,chang2024skews} and text-image generation tasks to further explore the casual factors of concept composition. We conduct controlled experiments on text-image datasets and investigate the influence of crucial factors: model size, dataset scale, and data distribution.

\begin{figure}[t]
  \centering
   \includegraphics[width=0.8\linewidth]{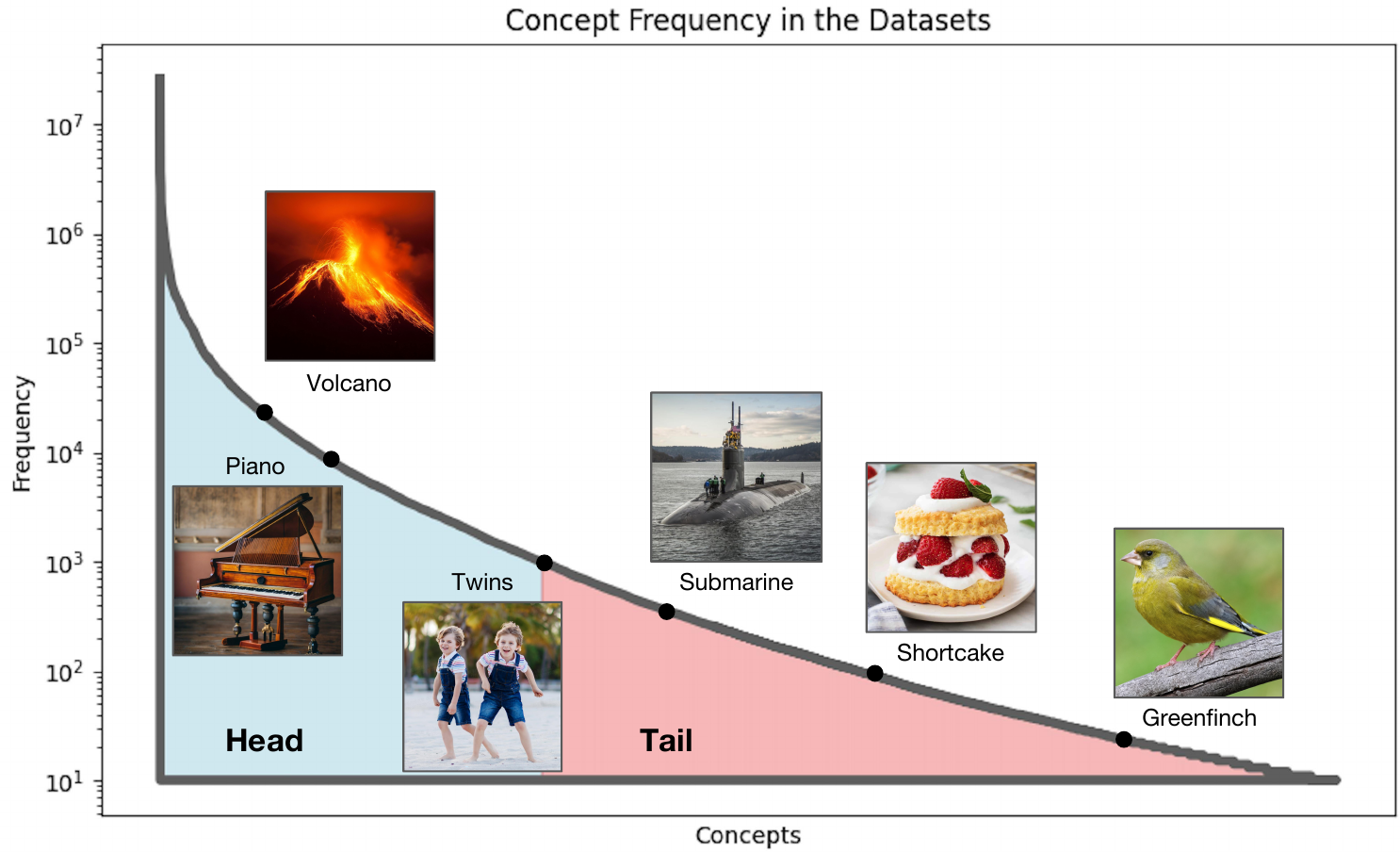}
   \vspace{-0.5em}
   \caption{Concept distribution of the datasets, which follows long-tail distribution.}
   \label{fig:data_distribution}
   \vspace{-1.5em}
\end{figure}

\textbf{Experiment Setup} To ensure dataset the same and training efficiency, we re-caption open-source datasets~\citep{laion} and collect a higher quality dataset, as existing open-source pre-trained models~\citep{midjourney,sd3,dalle3} do not provide their training data. Our dataset contains 31M text-image pairs with an average text length of 100 words and 20K noun concepts with distributions shown in Figure~\ref{fig:data_distribution}. To ensure the generality of the conclusion, we employ a DiT-based diffusion model~\citep{dit} without any special design. To accurately evaluate the model's concept composition ability rather than clarity, we introduce VQA based on VLM~\citep{gpt4v} for success rate as a quantitative metric except the CLIP-Score~\citep{clip}. We detect all noun concepts in the caption and pair them, requiring the VLM to verify the existence and relationships of concepts with the following questions: \textit{"Is concept A present in the image?"}, \textit{"Is concept B present in the image?"} and \textit{"Does the relationship between concepts A and B match the caption?"}. When all responses are "yes," the sample is considered successful.


\begin{figure}[t]
  \centering
   \includegraphics[width=0.8\linewidth]{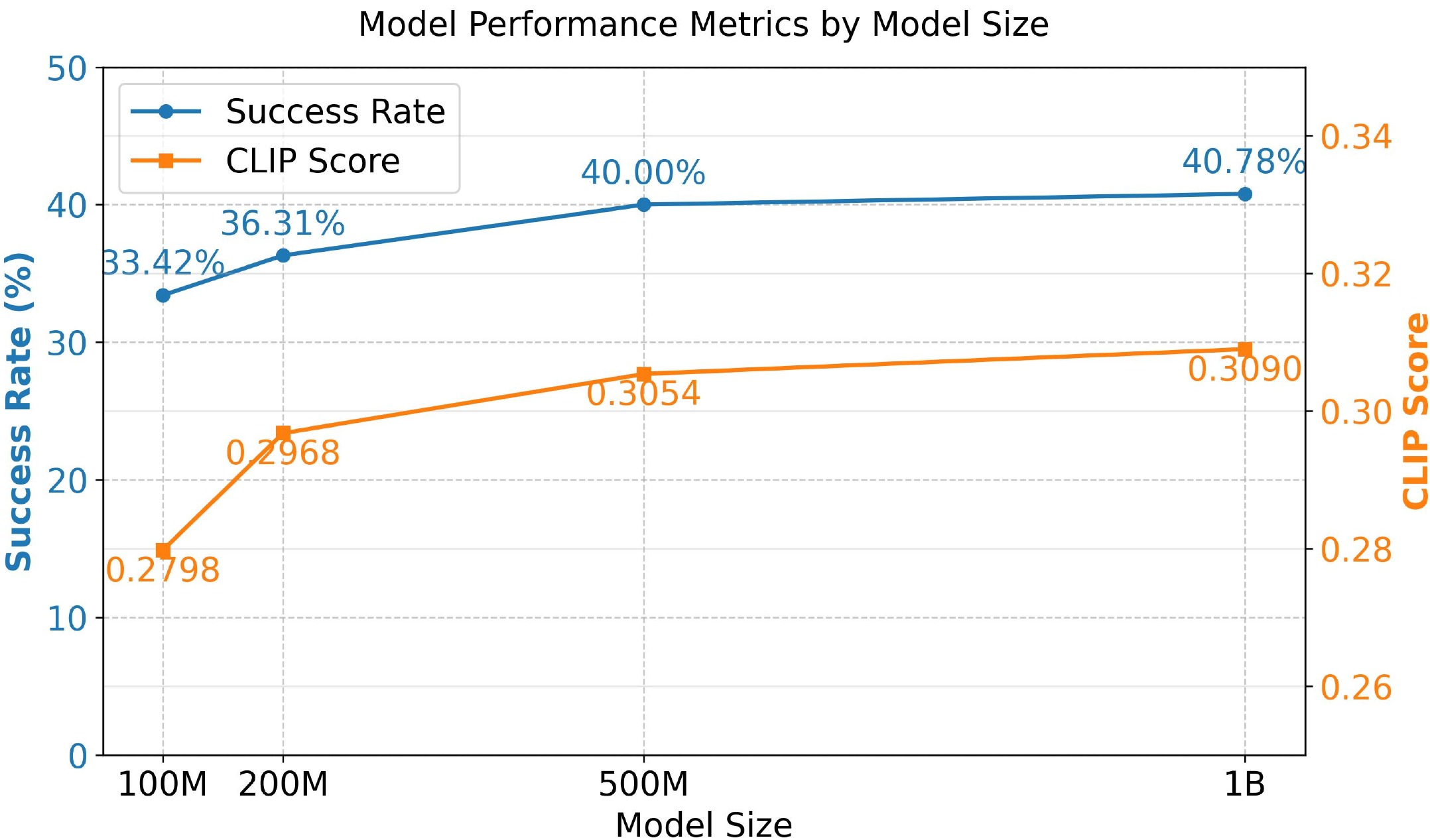}
   \caption{The performance of models with different parameter sizes under the LC-Mis benchmark~\citep{zhao2024lost}.}
   \label{fig:model_size}
   \vspace{-1.5em}
\end{figure}

\textbf{Model Size} In generation tasks, model size also follows the scaling law~\citep{kaplan2020scaling}, where models with more parameters have stronger generation ability. We keep the VAE from Stable Diffusion~\citep{sd3} the same and train diffusion models with 100M, 200M, 500M and 1B parameters on the same 31M dataset from scratch, differing only in the number of blocks and channels. We evaluate the CLIP score and success rate on the LC-Mis benchmark, as shown in Figure~\ref{fig:model_size}. When the model size exceeds 200M, the concept composition ability increases at a much slower rate as the model size grows.It indicates that model size is no longer the casual factor for concept composition when it reaches a certain magnitude relative to the dataset.

\begin{figure}[t]
  \centering
   \includegraphics[width=1.0\linewidth]{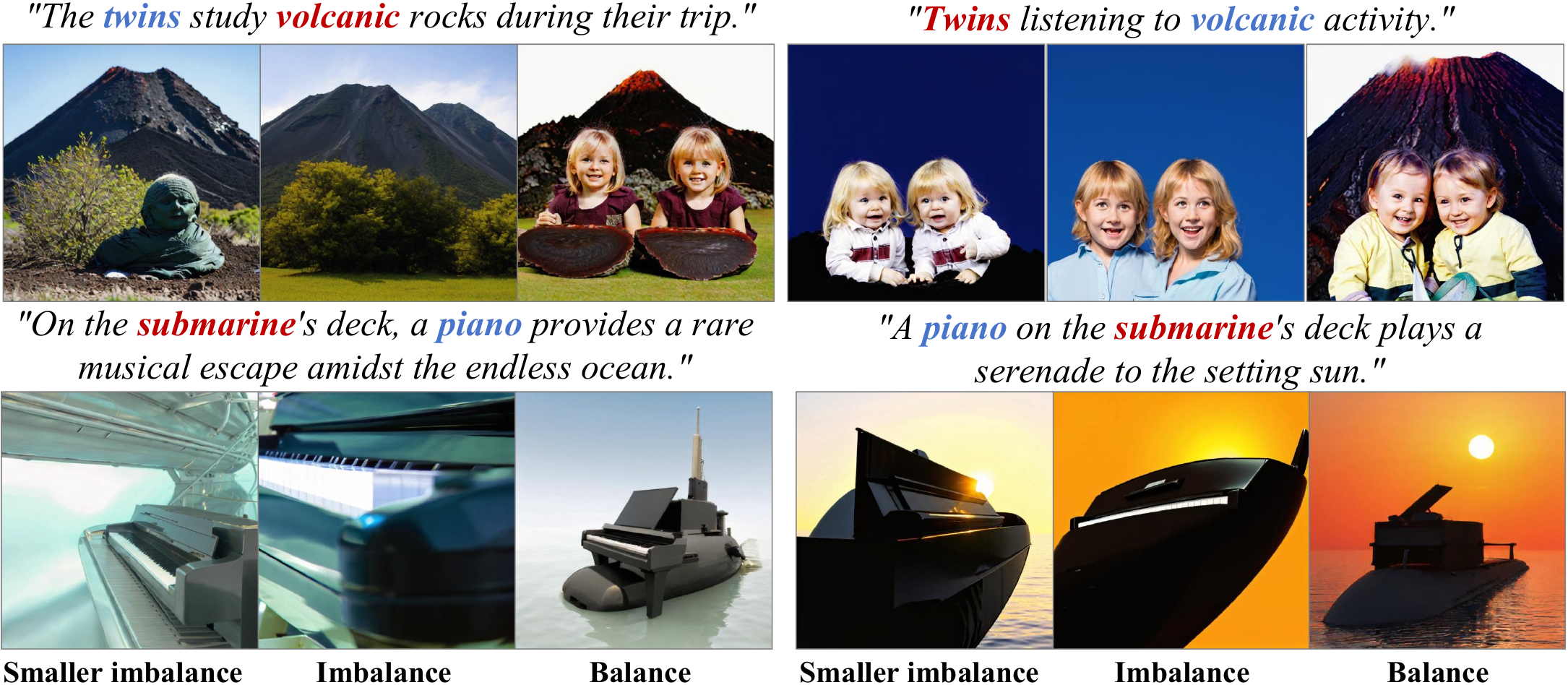}
   \vspace{-1em}
   \caption{The performance of models with different data scales and distributions.}
   \label{fig:analysis_comparison}
   \vspace{-1em}
\end{figure}

\textbf{Dataset Scale} It need to be clarified that the dataset scale here only refers to the number of samples for the same concept, not the co-occurrence of different concepts or data coverage, which we attribute to data distribution. To control for the same data distribution, we select two concepts from the dataset that have never co-occurred to form a pair, and extract the data containing them to create a new dataset with an imbalanced distribution. Then we maintain the sample ratio of the two concepts and sample two subsets with different scales from the new dataset, approximately following the same distribution. Finally, we resume the model which has never seen these concepts before, and conduct supervised finetuning on the two subsets for 20K steps. We generate 25 captions for each concept pair for VQA evaluation. To eliminate randomness, we select two pairs of concept composition ("piano-submarine" and "volcano-twins"), which follow imbalanced distributions of 100:1 and 10:1 respectively. As shown in Figure~\ref{fig:analysis_comparison}, despite the dataset being five times larger, there is no significantly improvement on the composition ability. And the number of failure cases does not decrease with the increasing dataset scale in Table~\ref{tab:analysis_metrics}. Therefore, simply increasing the data scale without changing the data distribution does not enhance the model's ability for concept combination.

\textbf{Data Distribution} We artificially construct balanced and imbalanced datasets for the experiment. We select two concepts from the dataset that have never co-occurred to form a pair, and extract the data containing them to create a new dataset with an imbalanced distribution, serving as the imbalanced dataset. Then, we downsample the data samples of head concepts to create a balanced dataset. We also use "piano-submarine" and "twins-volcano" as concept pairs for comparison, with sample ratios of 100:1 and 10:1 in the imbalanced dataset, and 1:1 in the balanced dataset. As shown in Figure~\ref{fig:analysis_comparison} and Table~\ref{tab:analysis_metrics}, although the imbalanced dataset contains all the data from the balanced dataset, the model trained on the balanced dataset still have stronger concept composition ability. Balanced data distribution can significantly boost the model’s ability to respond to combined concepts. Additionally, when most datasets in the open-world follow a long-tail distribution, addressing the impact of data imbalance is a crucial task for improving concept composition ability.

\begin{table}[]
\centering
\scriptsize
\resizebox{\linewidth}{!}{%
    \begin{tabular}{c|ccc|ccc}
    \toprule
    Pairs        & \multicolumn{3}{c|}{piano-submarine} & \multicolumn{3}{c}{volcano-twins} \\
    \midrule
    Head samples & 3K        & 15K         & 0.15K        & 1K        & 5K        & 0.5K       \\
    Tail samples & 0.03K       & 0.15K       & 0.15K       & 0.1K       & 0.5K      & 0.5K      \\
    \midrule
    Success rate &    16\%    &    20\%    &   56\%    &   28\%     &   20\%    &     64\%     \\
    CLIP Score   &    0.3076  &   0.3110   & 0.3226    &   0.2986   &  0.2948   &    0.3137   \\
    \bottomrule
    \end{tabular}
    }
\vspace{-0.5em}
\caption{The performance of models with different data scales and distributions.}
\label{tab:analysis_metrics}
\vspace{-2em}
\end{table}

\section{Method}
\label{sec:method}
\vspace{-0.5em}
We propose an online concept-wise equalization training strategy for data balancing, ensuring both effectiveness and efficiency. We first derive the form of the loss weight from the ideal data distribution in Section~\ref{subsec: theoretical}. Then we propose IMBA distance as a more accurate and efficient measure of data distribution in Section~\ref{subsec: IMBA distance}. Further, we introduce the novel online token-wise IMBA loss in Section~\ref{subsec: IMBA loss}. Finally, we extract inert concepts from open-set datasets to construct the new Inert-CompBench in Section~\ref{subsec: inert-bench}.

\subsection{Theoretical Analysis}
\label{subsec: theoretical}

Without loss of generality, we derive with the $\epsilon$-prediction DDPM~\citep{ddpm} framework. It is worth noting that this approach can be easily extended to various other diffusion model variants, such as flow matching~\citep{lipman2022flow}. In common implementations~\citep{dhariwal2021diffusion,song2020score}, the training loss of diffusion models parameterized by $\theta$ can be written as follow:
\begin{equation}
  L  = \mathbb{E}_{t,x_0,\epsilon } \left \| \epsilon -\epsilon _{\theta } \left (  x_t,y,t\right )  \right \| ^2 ,
  \label{eq:diffusion loss}
\end{equation}
where $y$ is the text prompts, $x_t$ is the noisy image at timestep $t$ from image $x_0$ and random Gaussian noises $\epsilon$. Inspired by existing work~\citep{somepalli2023diffusionreplication} where image regions respond to prompts as concept phrases, we rewrite the loss function as follow:
\begin{equation}
  \hat{L}  =\frac{\sum_{i=1}^{n}\mathbb{E}_{t,\epsilon }  \left \| \epsilon -\epsilon _{\theta } \left (  a_t^i,c_i,t\right )  \right \|^2 \varphi \left ( c_i \right )}{\sum_{i=1}^{n} \varphi \left ( c_i \right )},
  \label{eq:weight diffusion loss}
\end{equation}
where $c_i$ is a concept within the dataset containing $n$ concepts, $a^i$ is all image regions belonging to concept $c_i$, and $\varphi (c_i)$ is the frequency proportion of concept $c_i$ among all concepts. Naturally, we have $\sum_{i=1}^{n} \varphi(c_i)=1$, and assume that the set of image regions $A$ and the set of images $X$ have a one-to-one and onto mapping.

We assume that the optimal balance distribution of concepts is $\varphi^{*}(c_i)\sim U(1,n)$, which follows a discrete uniform distribution. Then we are able to write the optimal loss function based on Equation~\ref{eq:weight diffusion loss} as follow:
\begin{equation}
    \begin{aligned}
    \hat{L}^{*} & =\sum_{i=1}^{n} \mathbb{E}_{t,\epsilon }\left\|\epsilon-\epsilon_{\theta}\left(a_{t}, c_i, t\right)\right\|^{2} \varphi^{*}(c_i) \\
    & =\sum_{i=1}^{n}\mathbb{E}_{t,\epsilon }\left\|\epsilon-\epsilon_{\theta}\left(a_{t}, c_i, t\right)\right\|^{2} \varphi(c_i) \cdot \frac{\varphi^{*}(c_i)}{\varphi(c_i)},
    \end{aligned} 
    \label{eq:optimal loss}
\end{equation}
where $\frac{\varphi^{*}(c_i)}{\varphi(c_i)}$ is the loss weight we wish to obtain. Since $\varphi^{*}(c_i)$ is a constant when it follows a discrete uniform distribution, we only need to estimate $\frac{1}{\varphi(c_i)}$.

On the other hand, we wish that the model responds with similar intensity to different concepts~\citep{chefer2023attend}. Thus, we have the response intensity represented by the difference between the conditional and unconditional distribution as follow:
\begin{equation}
\left \| \epsilon_{\theta}\left(a_{t}, c_i, t\right) -\epsilon_{\theta}\left(a_{t}, \phi, t\right) \right \| = f(a_t),
\label{eq:concept response}
\end{equation}
where the intensity depends only on $a_t$. Meanwhile, we have the formulation of conditional and unconditional distribution in classifier-free guidance~\citep{ho2022classifierfree} as follows:
\begin{equation}
\epsilon_{\theta}\left(a_{t}, c_i, t\right) = \nabla_{a_t}  \log p\left ( a_t \right )+\sigma_t \nabla_{a_t}  \log p\left ( c_i |a_t \right ),
\label{eq:epsilon open}
\end{equation}
\begin{equation}
\epsilon_{\theta}\left(a_{t}, \phi, t\right) = \nabla_{a_t}  \log p\left ( a_t \right )+\sigma_t \nabla_{a_t}  \log p\left ( \phi |a_t \right ).
\label{eq:epsilon open phi}
\end{equation}
During training, the unconditional distribution is trained on the weighted expectation of multiple concepts:

\begin{equation}
    \begin{aligned}
    \epsilon_{\theta}\left(a_{t}, \phi, t\right) &=\frac{ \sum_{i=1}^{n} \varphi(c_i) \left [ \nabla_{a_t}  \log p\left ( a_t \right ) + \sigma_t \nabla_{a_t}  \log p\left ( c_i |a_t \right ) \right ]}{\sum_{i=1}^{n} \varphi(c_i)} \\
    &=\nabla_{a_t}  \log p\left ( a_t \right )+\frac{ \sigma_t \sum_{i=1}^{n} \varphi(c_i)  \nabla_{a_t}  \log p\left ( c_i |a_t \right )}{\sum_{i=1}^{n} \varphi(c_i)}.
    \end{aligned}
  \label{eq:uncond distribution}
\end{equation}

Then we apply Equation~\ref{eq:epsilon open},~\ref{eq:epsilon open phi} and \ref{eq:uncond distribution} into Equation~\ref{eq:concept response}:

\begin{equation}
\begin{aligned}
D_j &= \left \| \epsilon_{\theta}\left(a_{t}, c_i, t\right) -\epsilon_{\theta}\left(a_{t}, \phi, t\right) \right \| \\
  &= \left \| \sigma_t \nabla_{a_t} \log p\left( c_j|a_t \right) - \sigma_t \nabla_{a_t} \log p\left( \phi |a_t \right) \right \| \\
  &= \left \| \sigma_t \nabla_{a_t} \log p\left( c_j|a_t \right) - \frac{\sigma_t \sum_{i=1}^{n} \varphi(c_i) \nabla_{a_t} \log p\left( c_i |a_t \right)}{\sum_{i=1}^{n} \varphi(c_i)} \right \| \\
  &= \left \| \frac{\sigma_t}{\sum_{i=1}^{n} \varphi(c_i)} \left[ \nabla_{a_t} \log p\left( c_j|a_t \right) \sum_{i=1,i\ne j}^{n} \varphi(c_i) \right. \right. \\
  &\quad \left. \left. - \sum_{i=1,i\ne j}^{n} \varphi(c_i) \nabla_{a_t} \log p\left( c_i |a_t \right) \right] \right \| \propto \frac{1}{\varphi(c_j)} ,
\end{aligned}
\label{eq:IMBA distance concept}
\end{equation}

where $D_j$ is the difference between the conditional and unconditional distributions of concept $c_j$, and we refer it as \textbf{IMBA distance}. Since it is positively correlated with $\frac{1}{\varphi(c_i)}$, we use it to represent the frequency proportion of the concept in the dataset. Then we apply Equation~\ref{eq:IMBA distance concept} into Equation~\ref{eq:optimal loss} to obtain optimal loss:
\begin{equation}
    \begin{aligned}
    \hat{L}^{*} & =\sum_{i=1}^{n}\mathbb{E}_{t,\epsilon }\left\|\epsilon-\epsilon_{\theta}\left(a_{t}, c_i, t\right)\right\|^{2} \varphi(c_i) \cdot \frac{\varphi^{*}(c_i)}{\varphi(c_i)}\\
    & =\sum_{i=1}^{n}\mathbb{E}_{t,\epsilon }\left\|\epsilon-\epsilon_{\theta}\left(a_{t}, c_i, t\right)\right\|^{2} \\
    &\quad \left \| \epsilon_{\theta}\left(a_{t}, c_i, t\right) -\epsilon_{\theta}\left(a_{t}, \phi, t\right) \right \|_{sg}^\gamma \varphi(c_i) \\
    & \doteq \sum_{i=1}^{n}\mathbb{E}_{t,\epsilon }\left\|\epsilon-\epsilon_{\theta}\left(a_{t}, c_i, t\right)\right\|^{2} \left \| \epsilon -\epsilon_{\theta}\left(a_{t}, \phi, t\right) \right \|_{sg}^\gamma \varphi(c_i),
    \end{aligned} 
    \label{eq:loss weight concept}
\end{equation}
where we replace conditional distribution $\epsilon_\theta$ with ground truth $\epsilon$ for training stability. Finally, we rewrite loss function back to image $x$ and text prompts $y$:
\begin{equation}
  D  = \left \| \epsilon -\epsilon_{\theta}\left(x_{t}, \phi, t\right) \right \|_{sg}^\gamma,
  \label{eq:IMBA distance}
\end{equation}
\begin{equation}
  L^*  = \mathbb{E}_{t,x_0,\epsilon } D\left \| \epsilon -\epsilon _{\theta } \left (  x_t,y,t\right )  \right \| ^2 ,
  \label{eq:IMBA loss}
\end{equation}
where we implement the distance with $L-\gamma$ norm, and stopping gradient during training. We refer the novel loss function as \textbf{IMBA loss}. IMBA loss adapts dynamically during the training process without additional off-line modeling, ensuring both effectiveness and efficiency.

\subsection{IMBA Distance}
\label{subsec: IMBA distance}

Measuring data balance is particularly challenging due to the complexity of text prompts compared to classes~\citep{lin2017focal,qin2023classbalancing}. Additionally, with the exponential growth of datasets, offline data pruning will lead to data waste and significant computational and time costs. In this section, we demonstrate that the IMBA distance is able to represent the frequency proportion of concepts with both synthetic experiments and text-image generation experiments. We further give out the formulation of IMBA distance in Appendix~\ref{app:imba formulation}. 

\begin{figure}[t]
  \centering
   \includegraphics[width=1.0\linewidth]{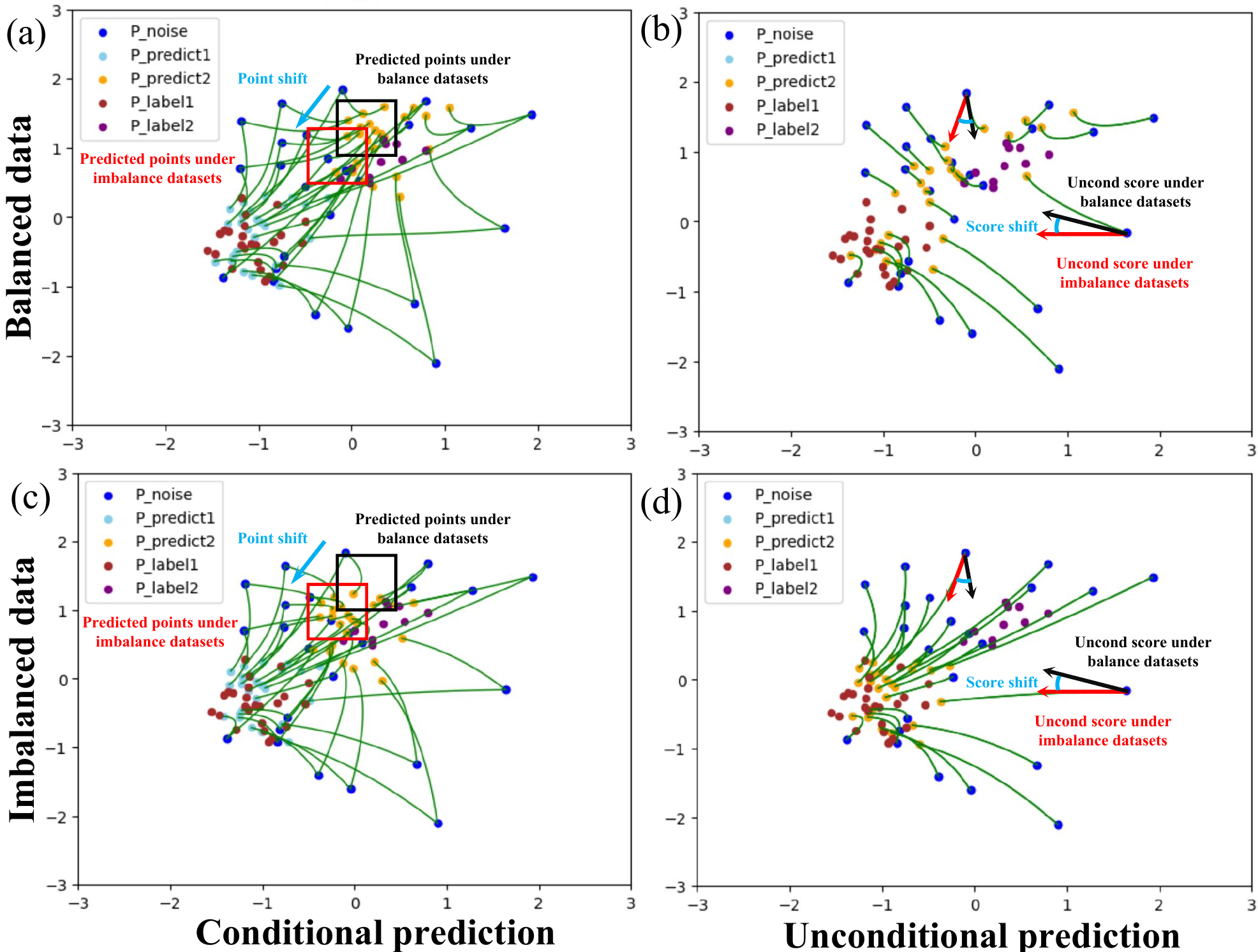}
   \caption{\textbf{Comparison between results from balance and imbalance datasets.} We simulate the training and inference results of diffusion models in a 2-dimensional space. With the dataset consisted by two classes (\textbf{\textcolor{brown}{brown}} and \textbf{\textcolor{violet}{purple}} points), diffusion models map random noise(\textbf{\textcolor{blue}{blue points}}) to the prediction(\textbf{\textcolor{orange}{yellow}} points) with flow matching(\textbf{\textcolor{teal}{green curve}}). Comparing Figure (a) and (c), imbalanced data leads to a \textbf{\textcolor{cyan}{shift}} from \textbf{black box} to \textbf{\textcolor{red}{red box}} on the prediction of tail concepts, harming the generalization of the tail concept(\textbf{\textcolor{violet}{purple points}}). Comparing Figure (b) and (d), imbalanced data makes unconditional score distribution tilt towards the head concept(\textbf{\textcolor{brown}{brown points}}) from \textbf{black arrow} to \textbf{\textcolor{red}{red arrow}}, proving that the difference between unconditional and conditional score distributions can serve as a metric for dataset distribution.}
   \label{fig:synthetic_experiments}
   \vspace{-1em}
\end{figure}

\subsubsection{Synthetic Experiments}
\label{subsubsec: synthetic experiments}
As shown in Figure~\ref{fig:synthetic_experiments}, we simulate the training and inference results of diffusion models in a 2-dimensional space. All data samples are represented by 2-dimensional coordinates and initialized into two classes, following normal distributions $P_1\sim U(-1,-0.3,0.1,0.1,0)$ (\textbf{\textcolor{brown}{brown points}}) and $P_2\sim U(0.3,1,0.1,0.1,0)$ (\textbf{\textcolor{violet}{purple points}}), respectively. We construct balanced and imbalanced training sets, both with 10K sample points in total, but with Class 1 (\textbf{\textcolor{brown}{brown points}}): Class 2 (\textbf{\textcolor{violet}{purple points}}) ratios of 1:1 and 99:1 respectively. Then we train a diffusion model with a 2-layer MLP on both datasets. During inference, random noise points (\textbf{\textcolor{blue}{blue points}}) following a standard normal distribution are mapped to the two target distributions through flow (\textbf{\textcolor{teal}{green curve}}) matching conditioned on the class, resulting in predicted points (\textbf{\textcolor{cyan}{sky blue}} and \textbf{\textcolor{orange}{yellow}}). In Figure~\ref{fig:synthetic_experiments}, figure (a,b) / (c,d) show the results of (conditional, unconditional) inference under balanced / imbalanced data respectively.

Comparing figures (a) and (c), we find that despite conditioning on the class during inference, the imbalanced data still pulls the prediction results (\textbf{\textcolor{orange}{yellow points}}) of the tail concept (\textbf{\textcolor{violet}{purple points}}) towards the head concept (\textbf{\textcolor{brown}{brown points}}). This is specifically shown by the predicted points in the \textbf{black box} in Fig.(a) drifting into the \textbf{\textcolor{red}{red box}} in Fig.(c) under the imbalanced data. This indicates that an imbalanced dataset harms the generalization of the tail concept. Meanwhile, comparing Fig.(b) and (d), we observe that under the balanced dataset, the unconditional score distribution points to the middle of the two classes (\textbf{black arrow} in Fig.(b)). But under the imbalanced dataset, it directly points to the head concept (\textbf{\textcolor{red}{red arrow}} in Fig.(d)), showing a very obvious score shift (\textbf{\textcolor{cyan}{blue curve}}). This demonstrates that the unconditional score distribution tends to favor the head concept under an imbalanced dataset, reducing the difference between the head concept's conditional distribution and unconditional distribution, consistent with the analysis in Equation~\ref{eq:IMBA distance}. Therefore, the IMBA distance can serve as a self-equilibrated, effective and efficient metric for data distribution during the training process.

\begin{table}[]
\scriptsize
\resizebox{\linewidth}{!}{%
    \begin{tabular}{c|cccc}
    \toprule
    Head concepts  & man  & girl & cat  & koala \\
    Frequency & 1.8M & 365K & 180K & 2.7K    \\
    \midrule
    Tail concepts  & gramophone & jeep & submarine & greenfinch \\
    Frequency & 756        & 420  & 332       & 24         \\
    \bottomrule
    \end{tabular}
    }
    \vspace{-0.5em}
    \caption{The frequency of concepts in the training set.}
    \label{tab:concept frequency}
\end{table}

\begin{figure}[t]
  \centering
   \includegraphics[width=1.0\linewidth]{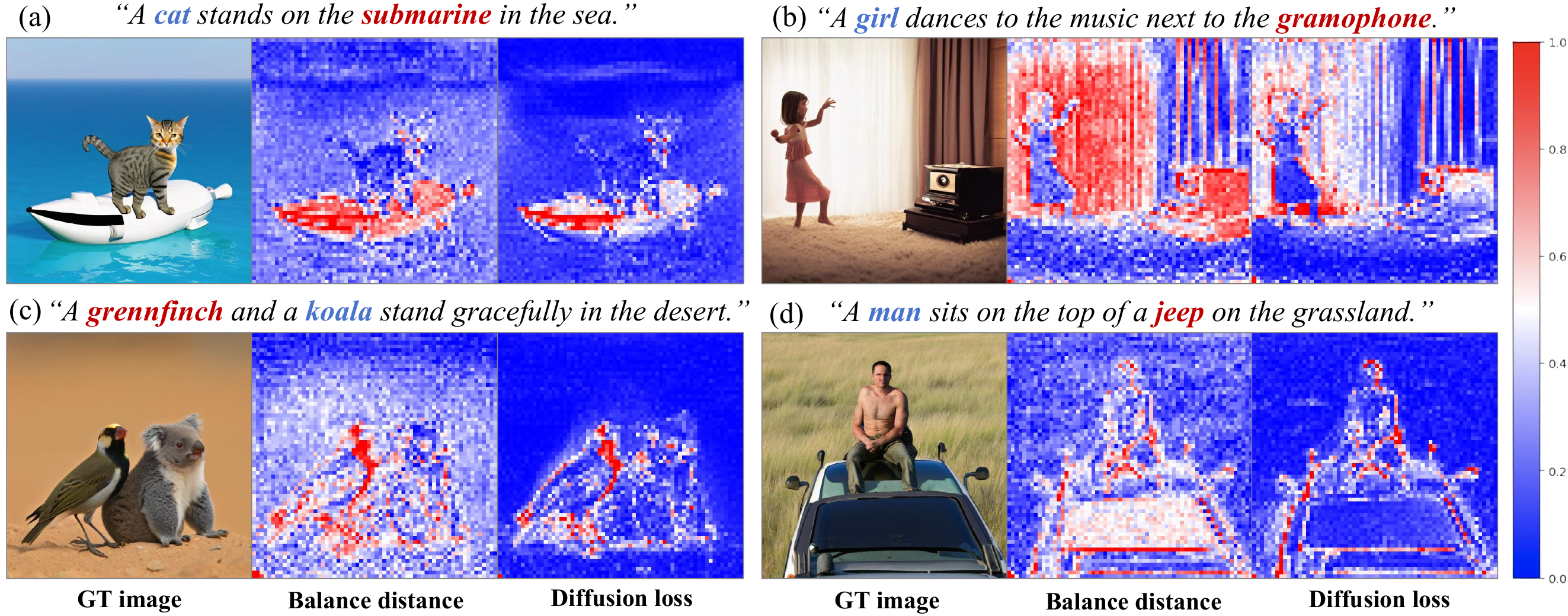}
   \vspace{-1em}
   \caption{\textbf{IMBA distance and diffusion loss of different concepts.} Tail concepts (\textbf{\textcolor{red}{red words}}) tend to have larger IMBA distance and diffusion loss than head concepts (\textbf{\textcolor{blue}{blue words}}).}
   \label{fig:text_image_experiments}
   \vspace{-1em}
\end{figure}

\subsubsection{Text-image Generation Experiments}
\label{subsubsec:text-image experiments}
In this section, we further demonstrate that IMBA loss can serve as the measure of data distribution on text-image experiments. We first collect a list of concepts from the training set with head and tail concepts each accounting for half in Table~\ref{tab:concept frequency}. Then, we pair head and tail concepts to create new captions to generate composition images. We select plausible samples to calculate IMBA distance and diffusion loss. In Equation~\ref{eq:IMBA distance}, since IMBA distance is related to $a_t$, it is not comparable under different content and timesteps. Therefore, we chose $t=1000$, when the entire image is random noise, to compare the IMBA distance of different concepts. We normalize both IMBA distance and diffusion loss to $[0,1]$, with the red area indicating larger values compared to the blue area. As shown in Figure~\ref{fig:text_image_experiments}, we find that tail concepts generally exhibit a larger IMBA distance compared to head concepts, which demonstrate that IMBA distance can measure the data distribution. Additionally, the diffusion loss is larger for tail concepts, indicating that they are under-fitting than head concepts. We further give out the formulation of IMBA distance in Appendix~\ref{app:imba formulation}.

\begin{algorithm}[t]
\caption{Training with IMBA loss}
\label{alg:IMBA loss}
\begin{algorithmic}[1]
   \Require Dataset $\{\mathbf{X},\mathbf{Y}\}$, noise scheduler $\bar{\alpha}$, model $\epsilon_\theta$
   \Repeat
   \State Sample data $(x_0, y)\sim\{\mathbf{X},\mathbf{Y}\}$
   \State Sample noise $\epsilon\sim\mathcal{N}(0,1)$ and time $t\sim\mathbf{U}(0,1)$
   \State Add noise with $x_t= \sqrt{\bar{\alpha}_t}x_0+ \sqrt{1-\bar{\alpha}_t}\epsilon$
   \State \textcolor{red}{Calculate weight $D=\left \| \epsilon -\epsilon_{\theta}\left(x_{t}, \phi, t\right) \right \|_{sg}^\gamma$}
   \State Conditional loss $L^*  = D\left \| \epsilon -\epsilon _{\theta } \left (  x_t,y,t\right )  \right \| ^2$
   \State Uncondtional loss $L_{u}  = \left \| \epsilon -\epsilon _{\theta } \left (  x_t,\phi,t\right )  \right \| ^2$
   \State Compute loss $L=\lambda L^*+(1-\lambda)L_u$
   \State Back propagation $\theta = \theta - \eta \nabla_\theta L$
   \Until{converged}
\end{algorithmic}
\end{algorithm}

\subsection{IMBA Loss}
\label{subsec: IMBA loss}
Here we present the process of our IMBA loss in Algorithm~\ref{alg:IMBA loss}. During training, after adding noise $\epsilon$ to the image $x_0$ to obtain $x_t$, we use the diffusion model $\theta$ to predict the conditional score $\epsilon _{\theta } \left (  x_t,y,t\right )$ and the unconditional score $\epsilon _{\theta } \left ( x_t,\phi,t\right ) $. The unconditional score and ground truth noise are used to calculate the IMBA distance $D$ based on the Equation~\ref{eq:IMBA distance} with stopping gradient. Then the IMBA distance weights the conditional loss from the conditional score and ground truth, resulting in the final IMBA loss $L^*$. Since the model also needs to train the unconditional distribution, we use the previously obtained unconditional score to calculate the loss, applying a weighting coefficient $\lambda$ to replace the original random drop condition. In the specific implementation, we set $\gamma=0.8$ to avoid color shifts and $\lambda=0.9$ which is the same as the mask ratio of the condition in the original diffusion training. The IMBA distance typically has a shape of (B, N, C), and we find that averaging over the channel dimension benefits training stability.

Compared to offline concept frequency-based loss weights~\citep{zhang2023deep,cui2019classloss,park2021influence}, IMBA loss is more accurate and efficient. Firstly, since text prompts are the joint distribution of multiple concepts, and each concept has a different frequency, it is challenging to intuitively and quickly set loss weights for samples as in class-related tasks. Secondly, while it is feasible to construct a concept graph where concepts are represented as nodes and the co-occurrence frequency of concept pairs as edge values to derive loss weights, the data distribution that the model learns evolves throughout the training process. This evolution creates a misalignment between the offline loss weights and the model's understanding. In contrast, IMBA distance is naturally coupled with the training process, providing a more consistent representation of data distribution. Thirdly, since different regions of an image may contain different concepts, data balancing should be performed at the concept region level. IMBA loss supports different loss weights for different regions, making it more accurate compared to a single weight for all regions. Additionally, with the exponential growth of data, constructing an offline concept graph becomes increasingly time-consuming and computationally intensive, making it challenging to reuse across different datasets. IMBA loss, on the other hand, only requires the calculation of IMBA distance with only a few lines of code during training, making it highly efficient and easily reusable. It is worth noting that our method is orthogonal to training-free methods and can be combined to further enhance the concept composition ability.

\begin{figure}[t]
  \centering
   \includegraphics[width=0.8\linewidth]{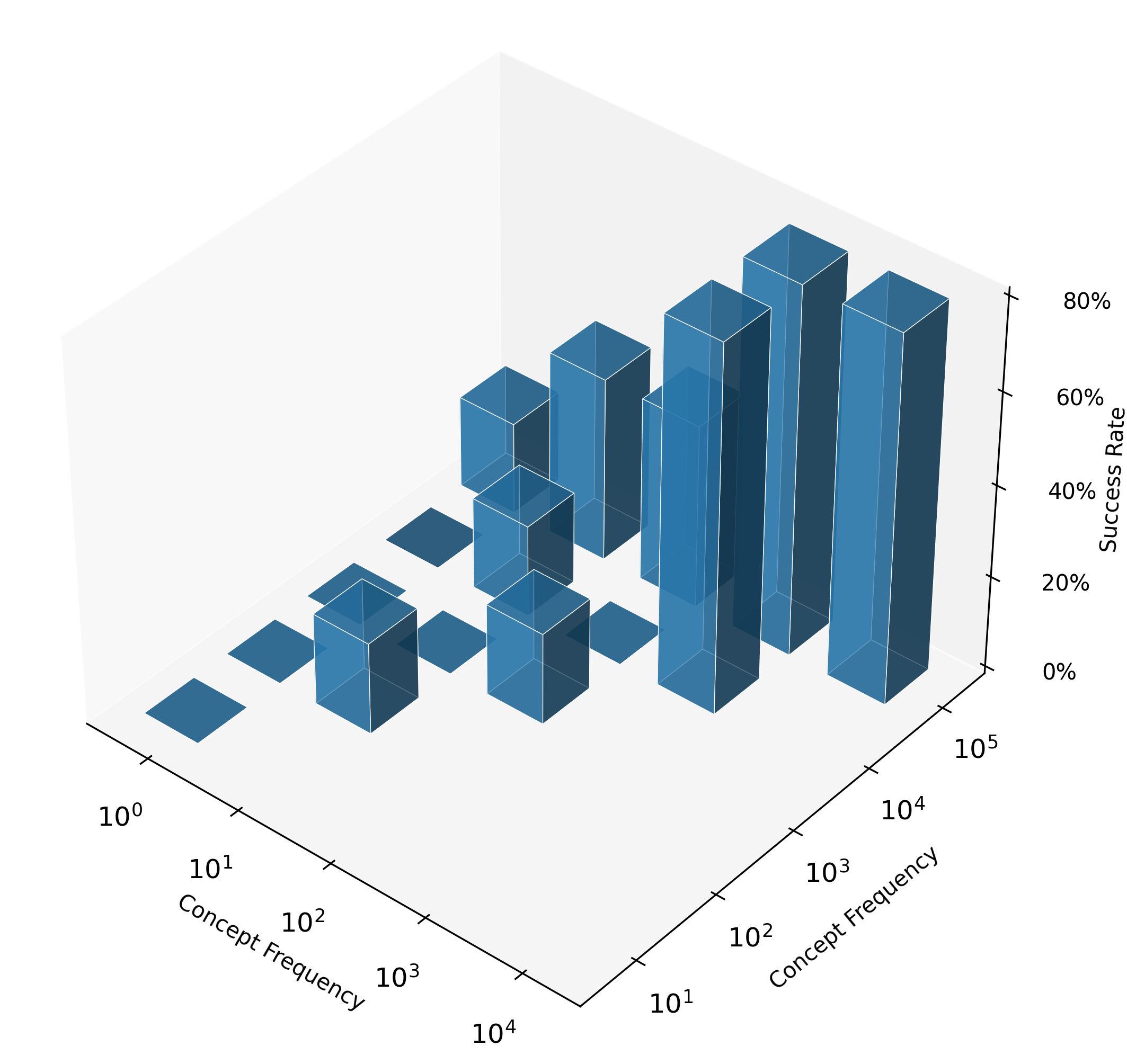}
   \caption{\textbf{Success rate of concepts with different frequency.} Tail concepts have lower success rate than head concepts.}
   \label{fig:tail_head_success_rate}
   \vspace{-1em}
\end{figure}

\subsection{Inert-CompBench}
\label{subsec: inert-bench}

Based on the analysis in Section~\ref{sec: analysis}, we further discover that some low-frequency concepts in datasets are difficult to successfully composite with other concepts, which we refer to as inert concepts. We first calculate the frequency of all noun concepts in the dataset, and sample 6 concepts uniformly from each frequency interval after taking the logarithm. Then we combine them in pairs to obtain 15 pairs and generate 5 captions for each pair with 75 captions in total. We evaluate the failure rate on the baseline models in Figure~\ref{fig:tail_head_success_rate}. We find that the success rate of concept composition increases as the concept frequency increases, indicating that tail concepts are more prone to failure cases compared to head concepts. Therefore, we should place more emphasis on tail concepts when constructing benchmarks. 

Due to the lack of attention to these concepts in existing benchmarks~\citep{huang2023t2icompbench,zhao2024lost}, we extract inert concepts from open-world datasets to construct a new benchmark called Inert-CompBench as a supplement. As shown in Algorithm~\ref{alg:compbench}, our framework comprises five phases: (1) Extract candidate sets where head concepts exhibit occurrence frequencies exceeding tail concepts by 100:1 ratio based on large-scale dataset statistics. (2) Select n domain-representative entity concepts from each pool through semantic typicality analysis. (3) Construct n×n Cartesian product combination space. (4) Build concept co-occurrence graphs to filter Top-K pairs with minimal structural associations, ensuring test cases reflect non-trivial compositional relationships. (5) Generate 5 linguistically diverse prompts per selected pair using GPT-4, ultimately forming a benchmark with 1K fine-grained test instances. This design forces models to handle statistically weak concept combinations, effectively revealing their compositional reasoning limitations.

\begin{algorithm}[t]
\caption{Inert-CompBench Construction}
\label{alg:compbench}
\begin{algorithmic}[1]
   \Require Dataset $\mathcal{D}$, concept count $n$, combination size $k$
   \State Extract head concepts $\mathcal{H}$ and tail concepts $\mathcal{T}$ from $\mathcal{D}$ where $\text{freq}(\mathcal{H})/\text{freq}(\mathcal{T}) > 100$ \hfill \textit{(Phase 1)}
   \State Select $n$ representative entities: $\hat{\mathcal{H}} \subset \mathcal{H}, \hat{\mathcal{T}} \subset \mathcal{T}$ via semantic typicality analysis \hfill \textit{(Phase 2)}
   \State Generate Cartesian product space $\mathcal{C} = \hat{\mathcal{H}} \times \hat{\mathcal{T}}$ with $|\mathcal{C}| = n^2$ \hfill \textit{(Phase 3)}
   \State Build co-occurrence graph $G$ from $\mathcal{D}$, select $\mathcal{C}^* = \text{Top-}k$ pairs with minimal edge weights in $G$ \hfill \textit{(Phase 4)}
   \State Generate prompts $\mathcal{P} = \bigcup_{c \in \mathcal{C}^*} \text{GPT-4}(c, 5)$, yielding $|\mathcal{P}| = 1k$ \hfill \textit{(Phase 5)}
   \Ensure Benchmark set $\mathcal{B} = \{\mathcal{P}, \mathcal{C}^*\}$
\end{algorithmic}
\end{algorithm}

\section{Experiments}
\label{sec:experiments}

\textbf{Setup.} To keep the dataset the same and improve training efficiency, we recaption and filter open-source data~\citep{laion}, resulting in a higher quality dataset of 31M samples for training. We employ a 1B parameter DiT-based diffusion model~\citep{dit} without any special design as the pipeline. Text prompts are injected into the diffusion model using T5 model~\citep{raffel2020t5}, and $512\times512$ images are encoded into the latent space using the VAE from Stable Diffusion~\citep{sd3}. \\
\textbf{Baseline.} We train the diffusion model from scratch for 4 epochs on the new dataset using the original diffusion loss to establish a baseline. To demonstrate the effectiveness of our method, we train the same model for the same epochs on the same dataset with the proposed IMBA loss. Additionally, we implement the training-free method~\citep{chefer2023attend} on the baseline for comparison. Furthermore, we finetune the baseline model from 3 epochs for 1 additional epochs with the IMBA loss, demonstrating that our method also provides significant benefits during fine-tuning. \\
\textbf{Benchmark.} We evaluate the concept composition ability of models on T2I-CompBench~\citep{huang2023t2icompbench}, LC-Mis~\citep{zhao2024lost} and Inert-CompBench. We employ VQA based on VLM~\citep{gpt4v} for evaluation in LC-Mis and Inert-CompBench as described in experiments setup of Section~\ref{sec: analysis}.

\begin{figure*}[t]
  \centering
  \vspace{-1em}
   \includegraphics[width=1.0\textwidth]{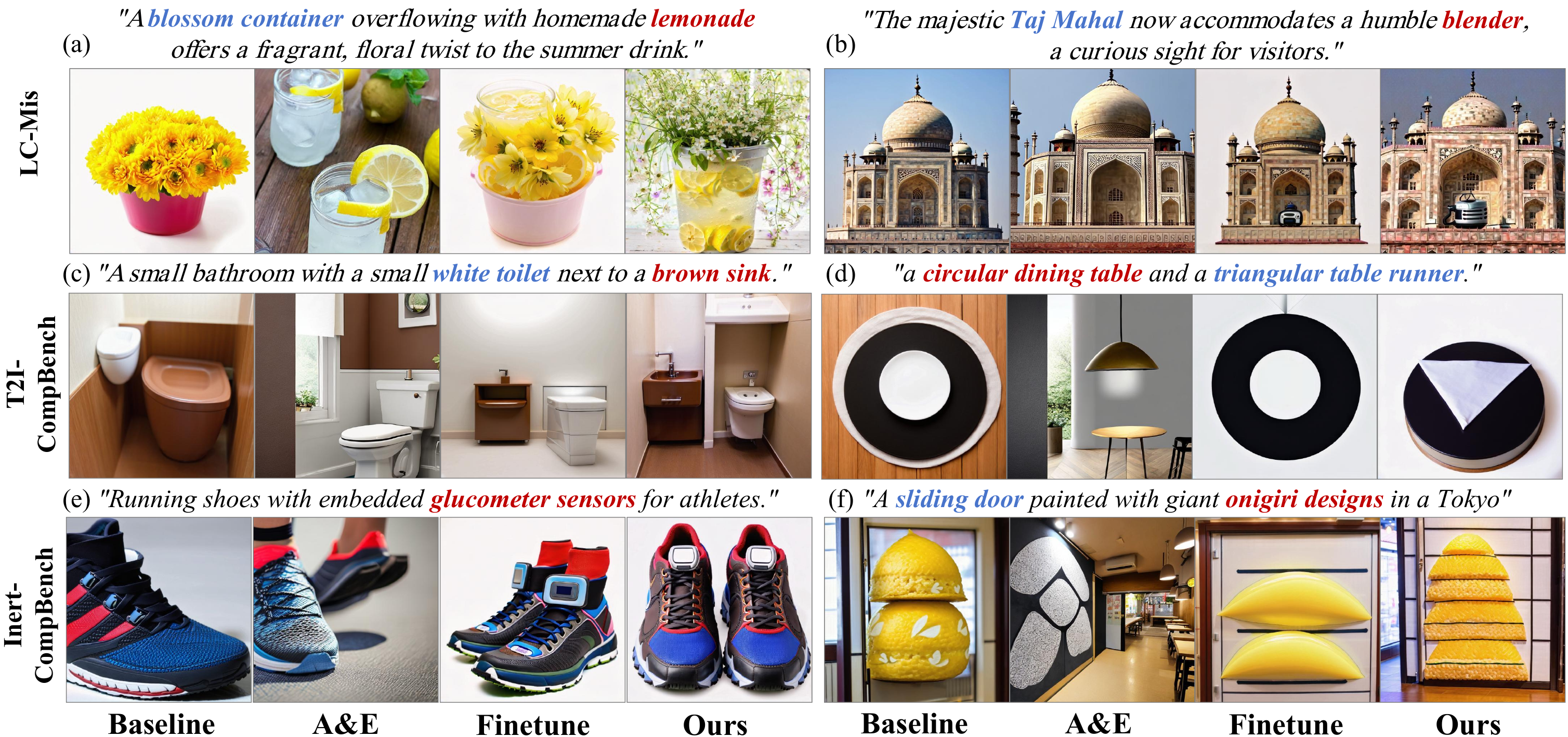}
   \caption{\textbf{Qualitative comparison with baseline.}}
   \label{fig:qualitative comparison}
   \vspace{-1em}
\end{figure*}

\begin{table*}[ht]
\tiny
\resizebox{\textwidth}{!}{
    \begin{tabular}{l|cc|ccccc|cc}
    \toprule
    \multirow{2}{*}{Model}   & \multicolumn{2}{c|}{LC-Mis~\citep{zhao2024lost}}      & \multicolumn{5}{c|}{T2I-CompBench~\citep{huang2023t2icompbench}}   & \multicolumn{2}{c}{Inert-CompBench}  \\
           & CLIP Score $\uparrow$ & VQA $\uparrow$ & Color $\uparrow$ & Shape $\uparrow$ & Texture $\uparrow$ & Non-spatial $\uparrow$ & Spatial $\uparrow$ & CLIP Score $\uparrow$ & VQA $\uparrow$ \\
    \midrule
    Baseline  & 0.3045  & 46.21\%   & 0.5812 & 0.4307 & 0.6188 & 0.3041& 0.1966& 0.3194  &  44\% \\
    A\&E~\citep{chefer2023attend} & \textbf{0.3198} & 48.42\%  &  0.6141   &  0.4378 &  0.6329  & \textbf{0.3078}  & 0.1998   & \textbf{0.3303}  & 44.5\%  \\
    Finetune   &  0.3073   &  \underline{51.82\%} & \underline{0.6668}  &   \underline{0.4919}   &  \underline{0.6575}  &   \underline{0.3075}   &   \underline{0.2218}    & 0.3172  &  \underline{46\%} \\
    Ours   & \underline{0.3121}     & \textbf{62.89\%}   & \textbf{0.7067} & \textbf{0.5151} & \textbf{0.6861} & 0.3071& \textbf{0.2518} &  \underline{0.3229} & \textbf{57\%} \\
    \bottomrule
    \end{tabular}
    }
\vspace{-1em}
\caption{\textbf{Quantitative comparison with baseline.} \textbf{Bold} font indicates the optimal value, and \underline{underlining} indicates the second-best value.}
\label{tab:quantitative}
\vspace{-1.5em}
\end{table*}

\subsection{Quantitative Comparison}
\label{subsec:quantitative}
As shown in Table~\ref{tab:quantitative}, we compare baseline, A\&E~\citep{chefer2023attend}, finetuning, and from-scratch training of our IMBA loss on LC-Mis~\citep{zhao2024lost}, T2i-CompBench~\citep{huang2023t2icompbench} and ours Inert-CompBench. Compared with the diffusion loss in the baseline, our IMBA loss can significantly improve the concept composition ability, whether training from scratch or finetuning from a pre-trained model. In addition, training from scratch yields better results compared to fine-tuning. Besides, when our approach achieves similar improvements in object missing, A\&E~\citep{chefer2023attend} performs slightly better on CLIP score, but far worse on attribute leakage (shape,color,texture,VQA). This is because A\&E is limited by the foundational generation model and cannot generate concepts that the generation model does not understand. Meanwhile, since the improvement of fine-tuning on Inert-CompBench is limited, it indicates that inert concepts require a longer training process to enhance concept composition ability.

\subsection{Qualitative Comparison}
\label{subsec:qualitative}
In Figure~\ref{fig:qualitative comparison}, we visualize the comparison results on three benchmarks separately, demonstrating the superiority of our method. We not only address object missing and attribute leakage effectively on existing benchmarks, but also demonstrate significant advantages under Inert-CompBench, greatly improving the success rate of inert concepts.

\textbf{Ablation.} We further conduct comprehensive experiments on sample-wise loss weight, hyper-parameters and IMBA distance. Please refer to Appendix~\ref{app:ablation}.

\vspace{-0.5em}
\section{Conclusion}
\label{sec:conclusion}
\vspace{-0.5em}
In this work, we propose a concept-wise equalization approach called IMBA loss for concept balancing to improve the concept composition ability of generation models. We first analyze the casual factor with elaborately designed experiments, bridging the gap between synthetic experiments and large-scale text-image generation. We demonstrate that data distribution has become the key factor when the model reaches a considerable size. Then, we propose the IMBA distance to estimate data distribution and demonstrate its effectiveness through both synthetic and text-image experiments. Subsequently, we introduce an online concept-wise equalization approach IMBA loss to balance concepts. Further, we identify inert concepts (difficult to integrate with other concepts) from large-scale text-image datasets and introduce the Inert-CompBench, complementing existing benchmarks. Finally, we conduct comprehensive experiments to demonstrate the priority of our methods.

\clearpage
{
    \small
    \bibliographystyle{ieeenat_fullname}
    \bibliography{main}
}

\newpage
\appendix
\maketitlesupplementary

\section{Formulation of IMBA Distance}
\label{app:imba formulation}
\begin{figure}[t]
  \centering
   \includegraphics[width=1.0\linewidth]{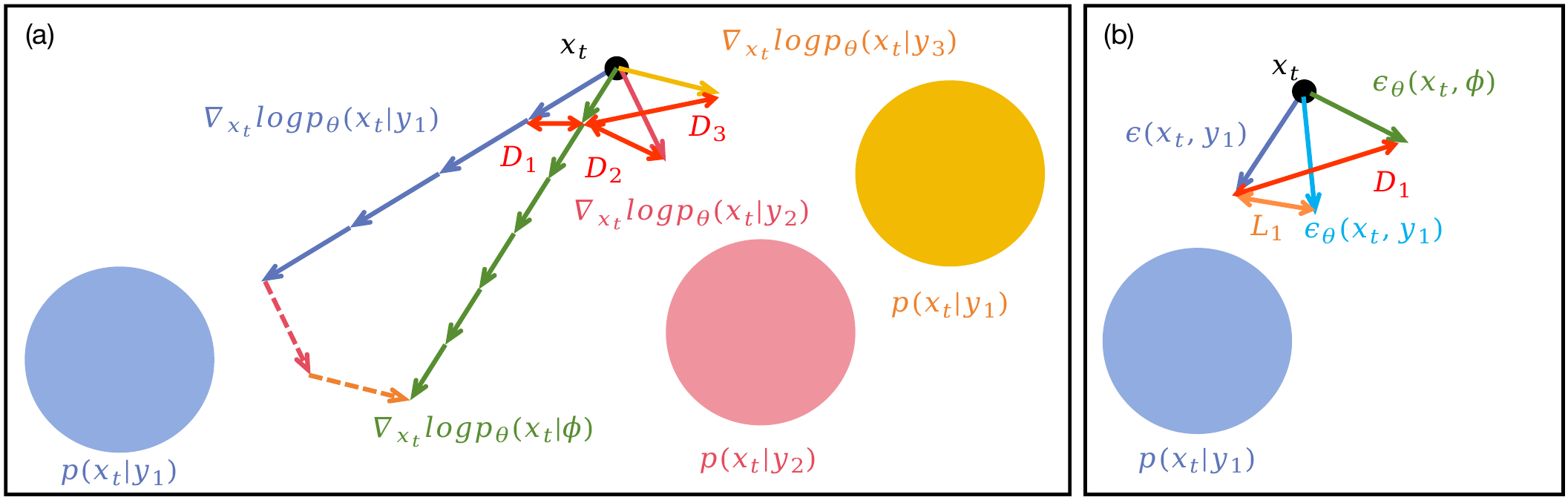}
   \caption{\textbf{Formulation of IMBA distance.} (a) \textcolor{teal}{Unconditional} distribution shifts toward \textcolor{blue}{head} concepts due to data imbalance, leading to a smaller IMBA distance \textcolor{red}{$D$} for head concepts. (b) Relationships between IMBA distance and diffusion loss during training.}
   \label{fig:imba_formulation}
   \vspace{-1em}
\end{figure}

Based on the above analysis, we can formulate the IMBA distance under the imbalanced data during training in Figure~\ref{fig:imba_formulation}. As shown in Figure (a), starting from the random noise $x_t$ in the latent space, conditional distribution points to different data distributions with different color based on different concepts. Due to data imbalance, concept $y_1$ has far more samples than other concepts. Since the unconditional distribution is weighted by all samples equally during training, it will shift toward concepts with more samples like the green arrow, leading to a smaller IMBA distance $D_1$. In the training set, the ratio of samples between head and tail concepts often reaches a factor of thousands, far exceeding the ratio shown in the figure, indicating a much more severe data imbalance issue and more pronounced pattern of IMBA distance. As shown in Figure (b), original diffusion loss represents the distance between the conditional distribution and the predicted conditional distribution, and IMBA distance represents the distance between the predicted conditional distribution and the unconditional distribution. Specifically, when IMBA distance is implemented with the L2 norm, it is equivalent to the unconditional loss.

\section{Ablation study}
\label{app:ablation}

\subsection{Stability of IMBA Distance}
\begin{figure}[t]
  \centering
   \includegraphics[width=1.0\linewidth]{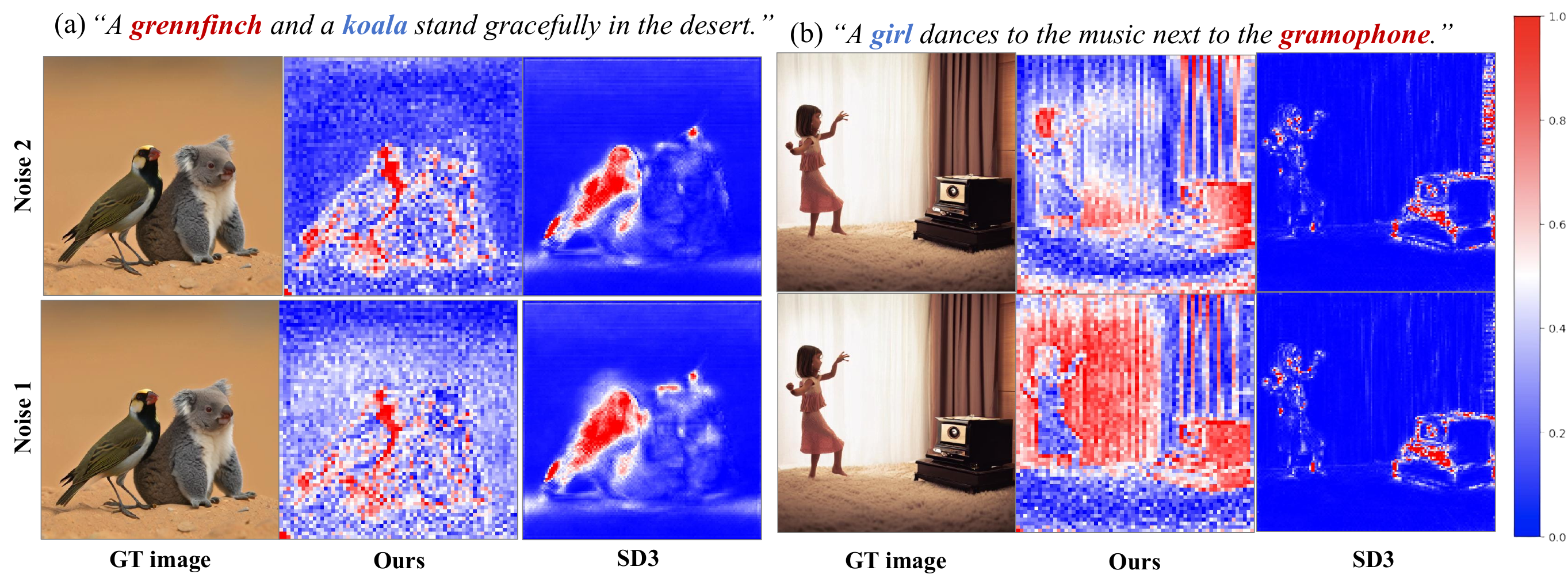}
   \vspace{-1em}
   \caption{\textbf{IMBA distance of different models and noises.}}
   \label{fig:imba_model_noise}
\end{figure}

In Figure~\ref{fig:imba_model_noise}, we calculate the IMBA distance of the same prompt on models with different size, architecture and noise. We find it is stable across all settings.

\subsection{Comparison with Frequecy-based Method}
 Since the text is a joint distribution of multiple concepts, it is difficult to calculate weights from a frequency perspective, and there is little concept-balancing work for text-to-image generation. Therefore, we compare IMBA loss with a frequency based method on class-image~\citep{cui2019classloss}. We sample 5 concepts each from the head and tail concepts and combine the data containing these concepts in the training set into a new subset. We then finetune the model on the subset using the frequency-based and our method respectively. Meanwhile, we pair the 10 concepts to generate 5 captions for each pair as the test set. As shown in the Table~\ref{tab:balancing_method_ablation}, our method outperforms the frequency-based method.

\begin{table}[]
\centering
\tiny
\vspace{-1em}
\resizebox{0.8\linewidth}{!}{%
    \begin{tabular}{ccccc}
    \toprule
    Loss weight & Baseline  & Frequency-based & Ours \\
    \midrule
    Success rate & 33.3\% & 49.3\% & 65.7\% \\
    CLIP Score & 0.3113 & 0.3101 & 0.3218  \\
    \bottomrule
    \end{tabular}
    }
\vspace{-1em}
\caption{The performance of different balancing methods.}
\label{tab:balancing_method_ablation}
\vspace{-2em}
\end{table}

\subsection{Comparison with Sample-wise Loss Weight} 
\begin{figure}[t]
  \centering
   \includegraphics[width=1.0\linewidth]{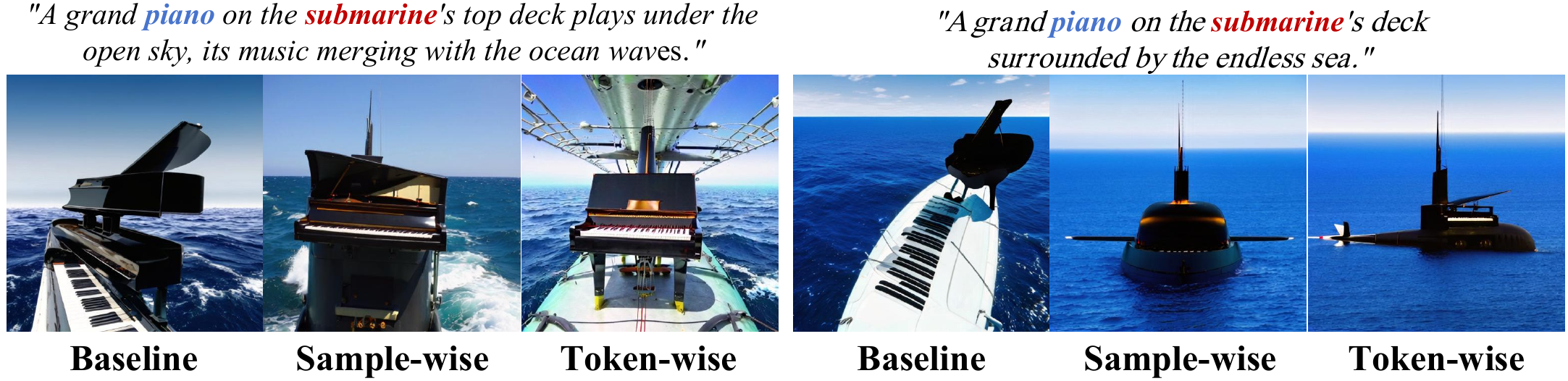}
   \vspace{-1.5em}
   \caption{The performance of models with different loss weight.}
   \label{fig:weight_ablation}
   \vspace{-1em}
\end{figure}

We finetune the same model on the imbalanced "piano-submarine" subset for 10K steps with sample-wise and our token-wise loss weight respectively. As shown in Table~\ref{tab:loss_weight_ablation} and Figure~\ref{fig:weight_ablation}, all results are evaluated on 25 captions. And sample-wise loss weight performs better than the baseline due to the reweight balancing. Meanwhile, token-wise loss weight achieves the best performance since it applies more fine-grained weights on different image regions according to concepts.

\begin{table}[]
\centering
\tiny
\resizebox{0.8\linewidth}{!}{%
    \begin{tabular}{ccccc}
    \toprule
    Loss weight & Baseline  & Sample-wise & Token-wise \\
    \midrule
    Success rate & 32\% & 64\% & 72\% \\
    CLIP Score & 0.2924 & 0.3022 & 0.3106  \\
    \bottomrule
    \end{tabular}
    }
\vspace{-0.5em}
\caption{The performance of models with different loss weight.}
\label{tab:loss_weight_ablation}
\vspace{-0.5em}
\end{table}

\begin{figure}[t]
  \centering
   \includegraphics[width=1.0\linewidth]{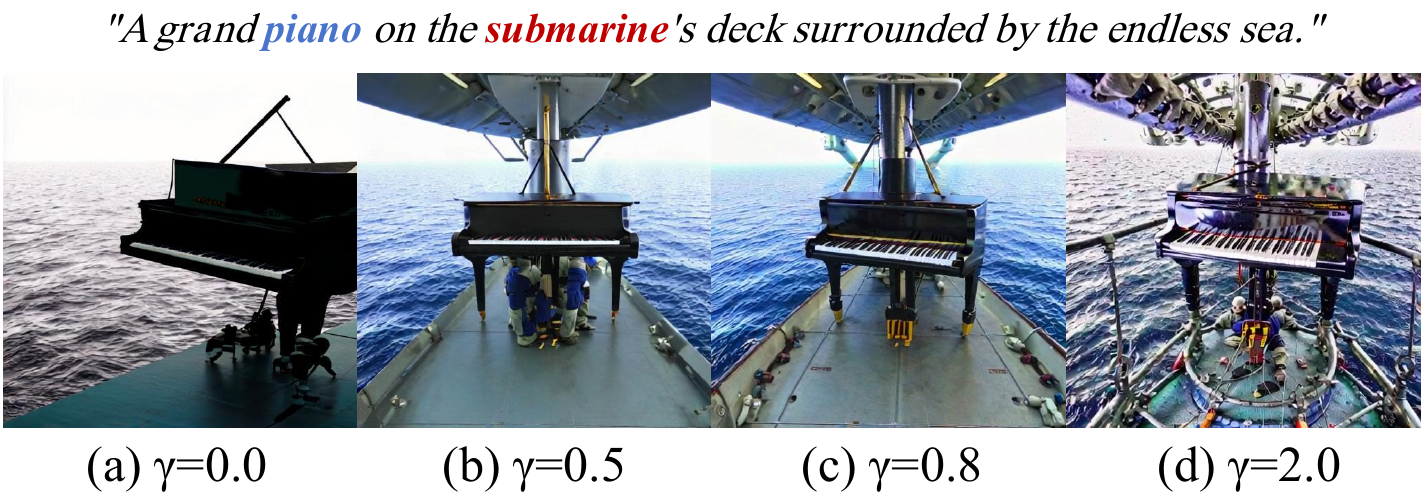}
   \caption{\textbf{Results from models trained with different $\gamma$.}}
   \label{fig:gamma_ablation}
   \vspace{-1em}
\end{figure}

\vspace{-0.5em}
\subsection{Hyper-parameters of IMBA Loss}
We train the model on the "piano-submarine" subset to conduct ablation experiments on the value of $\gamma$. Specifically, when $\gamma=0.0$, IMBA loss is equivalent to the original diffusion loss. When $\gamma=2.0$, the value of the IMBA distance equals the value of the unconditional loss. As shown in Figure~\ref{fig:gamma_ablation}, when $\gamma$ approaches 0.0, the concept composition ability of the model diminishes, as the semantic of the submarine in Figure(a) almost disappears. When $\gamma$ approaches 2.0, the model exhibits severe color shift issues as seen in Figure(b). We chose $\gamma=0.8$ based on these observations.

\subsection{IMBA Distance after Training.}
\label{app:visualizaiton of loss}

\begin{figure}[t]
  \centering
   \includegraphics[width=1.0\linewidth]{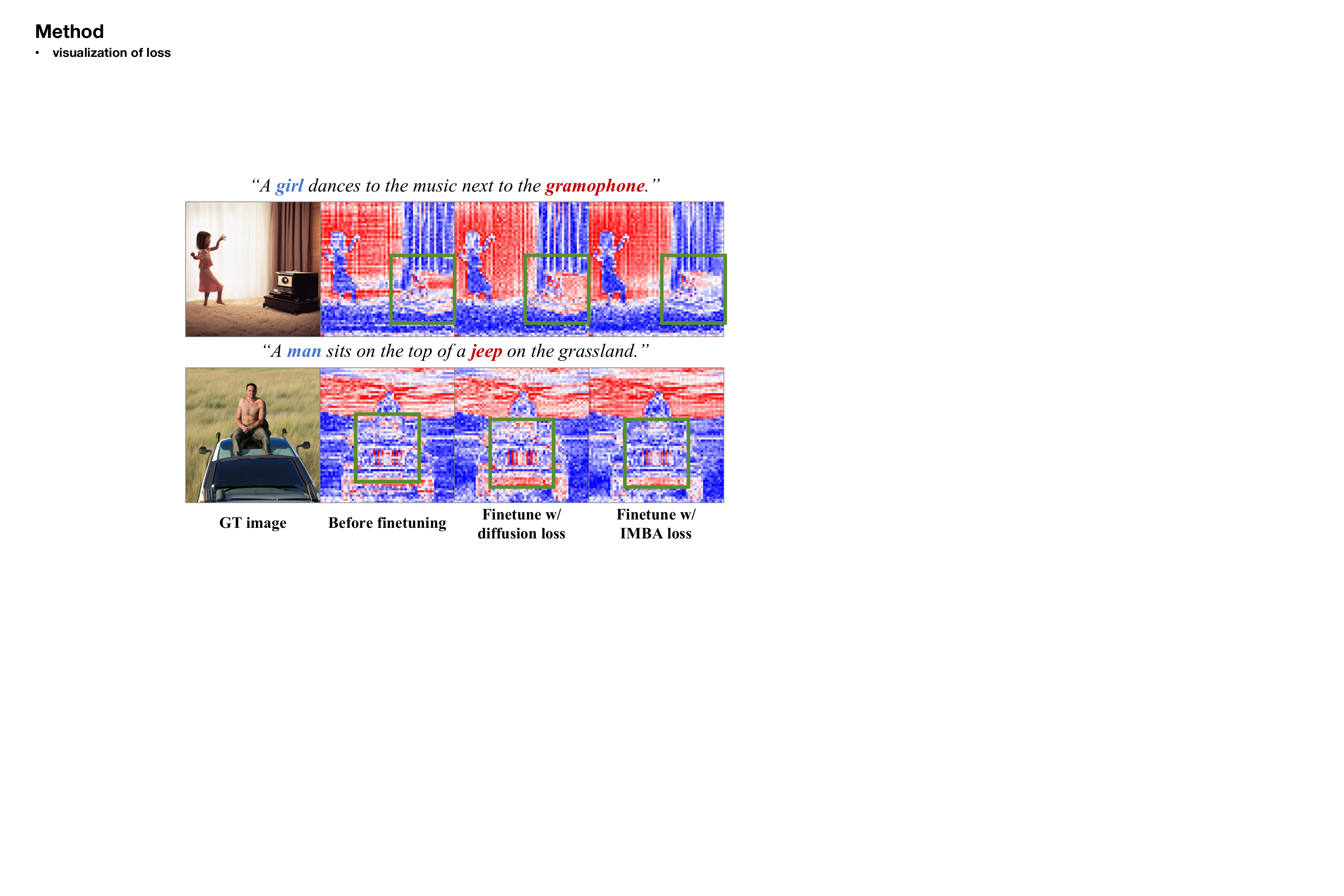}
   \vspace{-1.5em}
   \caption{IMBA distance before and after finetuning.}
   \label{fig:app_viz_distance}
   \vspace{-0.7em}
\end{figure}

We resumed training a model for 3 epochs using diffusion loss, and then fine-tuned it separately with diffusion loss and IMBA loss. The difference in IMBA distance between the two models after fine-tuning is shown in Figure~\ref{fig:app_viz_distance}. It can be observed that, due to concept balancing during the training process with IMBA loss, the IMBA distance after training with IMBA loss pays more attention to tail concepts (\textbf{\textcolor{red}{red words}}). Consequently, the IMBA distance in the corresponding regions (\textbf{\textcolor{green}{green boxes}}) is smaller compared to training with diffusion loss.

\vspace{-0.5em}
\section{More Experiment Results of the Model Size}
\label{app:model size}
\begin{figure}[t]
  \centering
   \includegraphics[width=0.7\linewidth]{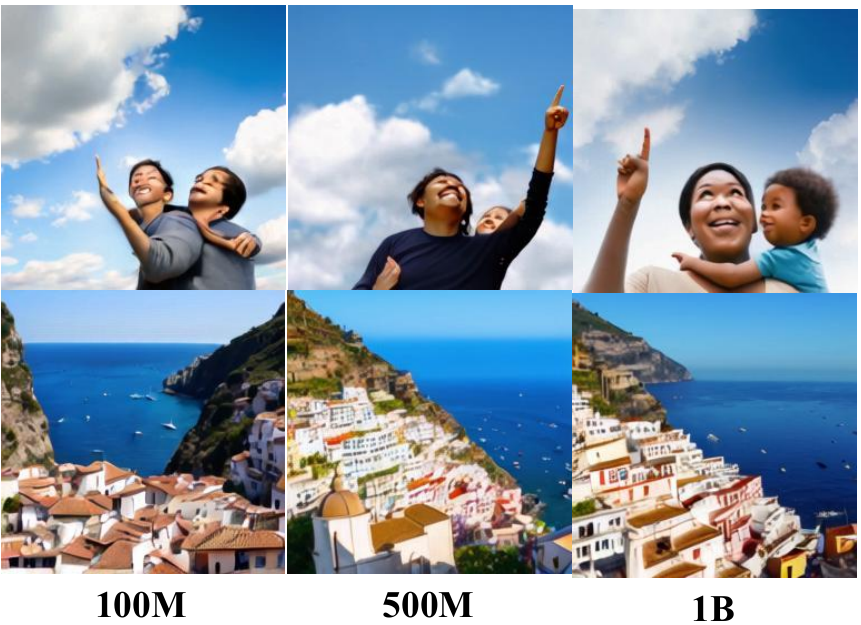}
   \vspace{-1em}
   \caption{Generation results of models with different sizes from the same initial noise.}
   \label{fig:app_semantic}
   \vspace{-1em}
\end{figure}

When testing different model sizes on the same dataset in Section~\ref{sec: analysis}, we observed that even with significant differences in model size, the generated images exhibit highly similar structural features given the same initial noise and text prompts, as illustrated in Figure~\ref{fig:app_semantic}. This suggests that once a model reaches a certain size, the dataset itself becomes more influential in determining the generated images rather than the model capacity. Larger models indeed have better convergence capabilities, but they do not dictate the high-dimensional semantics or concept composition abilities of the images.

\end{document}